\documentclass[10pt,twocolumn]{article}

\usepackage[letterpaper,margin=0.75in,columnsep=0.25in]{geometry}
\usepackage[T1]{fontenc}
\usepackage[utf8]{inputenc}
\usepackage{newtxtext,newtxmath}
\usepackage{microtype}

\usepackage{booktabs}
\usepackage{multirow}
\usepackage{amsmath}
\usepackage{graphicx}
\usepackage{tikz}
\usetikzlibrary{shapes.geometric,arrows.meta,positioning,calc,fit,backgrounds}
\usepackage[svgnames,dvipsnames]{xcolor}
\definecolor{exred}{RGB}{176,42,55}
\definecolor{gateblue}{RGB}{28,62,110}
\definecolor{gatefill}{RGB}{234,239,247}
\definecolor{exfill}{RGB}{252,240,240}
\usepackage{stfloats}
\usepackage[numbers]{natbib}
\usepackage{url}
\usepackage[hidelinks]{hyperref}
\usepackage{authblk}

\renewenvironment{abstract}{\par\noindent\textbf{\abstractname.}\ \ignorespaces}{\par\medskip}

\title{\textbf{Grading the Unannotated: Insights from the WorkRB Skill Extraction
  Challenge at RecSys-HR 2026}}

\author[1,3]{Warre Veys\thanks{Corresponding author.}}
\author[2]{Luis Gasco}
\author[1]{Jens-Joris Decorte}
\affil[1]{TechWolf, 9000 Gent, Belgium}
\affil[2]{Avature, Barcelona, Spain}
\affil[3]{Ghent University -- imec, 9052 Gent, Belgium}
\affil[ ]{\texttt{workrb@techwolf.ai}}
\date{}

\begin{document}

\maketitle

\begin{abstract}
Skill extraction benchmarks score models against a small set of human-annotated positives and treat
every other skill in the vocabulary as equally wrong. At the scale of ESCO (13,939 skills) this
binary view can rewards models that rank the annotated positive high for the wrong reason, and hides
predictions that are semantically nonsensical. The WorkRB Challenge at RecSys-HR 2026 replaces this
binary protocol with graded relevance: every candidate skill in a model's ranking window carries a 0
to 4 relevance grade, and the leaderboard scores nDCG@100 against these grades. Since ternary
judgements already proved too coarse for objective annotation, relevance is decomposed into a cascade
of four simple binary questions (same domain, activity mentioned, right granularity, replacement of
a gold positive) that a smaller LLM judge can answer reliably, and the resulting labels are validated with
blind human and frontier-model re-annotation. This paper documents the annotation methodology, the
construction of the five benchmark datasets, four with vacancy sentences as queries and one with
lexically distant skill surface forms, the validation of the LLM cascade, and the insights from the
challenge submissions. The field converged on multi-stage retrieve-and-rerank pipelines, whose
gains come from clearing nonsense out of the ranking window rather than from finding more
positives: at the top of the leaderboard, the graded ordering is uncorrelated with what binary
evaluation would have said. The graded labels expose substantially bigger gaps between the top
submissions and the baselines than the binary protocol does, and show that the domain of
semantic-understanding skill recommenders is still wide open. The clear direction after this
challenge is bringing efficiency back: every submitted system we could size spends at least $42\times$,
and typically several hundred times, more parameters than the baselines it beats.
\end{abstract}


\section{Introduction}

Skill recommenders decide who gets seen at work. The ranking a model returns for a vacancy sentence
or a profile feeds crucial applications like hiring shortlists, internal mobility and project staffing, 
promotion and succession pools, and the learning paths an organisation offers. A skill placed in the top of such
a ranking for the wrong reason does not simply cost a point of accuracy: it attaches a requirement
to a role that the role does not have, or attaches a capability to a person who was never assessed
on it, and both propagate into decisions about that person. Nonsense high in the window is
therefore not only a quality problem but a fairness one, and it is precisely the failure that the
prevailing evaluation protocol cannot see.

Previous skill extraction evaluation frameworks share a blind spot: every unannotated skill is
treated as equally wrong. With a vocabulary of 13,939 ESCO skills~\cite{esco2024} and only a handful
of human annotated positives per sentence, this assumption does not hold.
\citet{veys2026surface} showed that the standard training setup for skill recommenders
is prone to lexical shortcuts, potentially ranking skills by surface overlap with the input rather than by
meaning, and that binary ranking metrics hide this completely: the annotated positive and the
shortcut often coincide, so MAP stays high even when some of the other top predictions are nonsense
from an unrelated domain.

The remedy proposed there is graded evaluation: instead of positive versus everything-else,
distinguish predictions that are perfect, merely plausible, and clearly wrong. Their ternary rubric
(Positive, Acceptable, Nonsense) exposed the failure mode but proved too coarse as an annotation
target: the Acceptable bucket collapses near-misses, broad umbrella skills, and genuinely additional
correct skills into one class. Asking an annotator (human or LLM) for a direct 0 to 4 grade
invites inconsistency: assessor agreement is known to degrade as relevance scales grow
finer~\cite{voorhees2000variations,roitero2018fine}, something Section~\ref{sec:validation} measures the
same instability for our own judge.

This challenge tries to close that gap with an automatic annotation framework built for objectivity. 
We decompose the graded judgement into a tree of four binary questions, each simple
enough to optimise and answer reliably in isolation (Section~\ref{sec:rubric}). An LLM judge walks the cascade
for every (sentence, skill) candidate pair, producing dense graded relevance labels over the
top-2000 retrieval candidates per query. 

The challenge poses the task of ranking all of ESCO against a free-text query. It runs over five
datasets that differ only in what that query is. Four are existing, binary annotated, skill
extraction datasets whose queries are vacancy sentences, re-annotated with graded relevance. The
fifth targets skill normalisation: its queries are lexically distant surface forms of individual
ESCO skills, built from ESCO's own alternative labels filtered to be lexically far but semantically
equivalent (Section~\ref{sec:norm}), and it is annotated for full-vocabulary recommendation in
exactly the same way. All five are served through WorkRB, an open benchmark library that normalizes 
evaluation of AI in the work domain~\cite{delange2026workrb}, and the
challenge allows any technique that produces rankings.

\paragraph{Contributions.} We (1) introduce a binary-cascade annotation framework that turns graded
relevance into a sequence of objective yes/no decisions, (2) release graded validation and test
annotations for five benchmarks, four established skill extraction datasets and a new
skill normalisation dataset, all annotated for the same full-vocabulary recommendation task, (3)
validate the LLM-produced labels with blind human and frontier-model re-annotation, and (4)
report what the challenge's 154 valid submissions reveal: a field that converged on multi-stage
retrieve-and-rerank pipelines at least $42\times$ larger than the trained baselines, and a top of the
leaderboard whose ordering binary evaluation cannot reproduce (Kendall $\tau = 0.06$ against binary
MAP over the twelve best teams).

Read together, these results say something the binary leaderboards of the last few years did not.
Semantic skill understanding was not a solved problem going into this challenge: under graded
evaluation the strongest, efficient, published skill extractor available off the shelf sits $0.24$ nDCG@100 below
what a field of motivated participants could reach, a gap the binary protocol reports as a third of
that size. It is also not solved coming out of it. What closed the gap was scale and pipeline depth,
not a better skill encoder, and the road to an open-source semantic-understanding skill recommender that an organisation
could actually deploy, remains open.

\section{The graded relevance rubric}\label{sec:rubric}

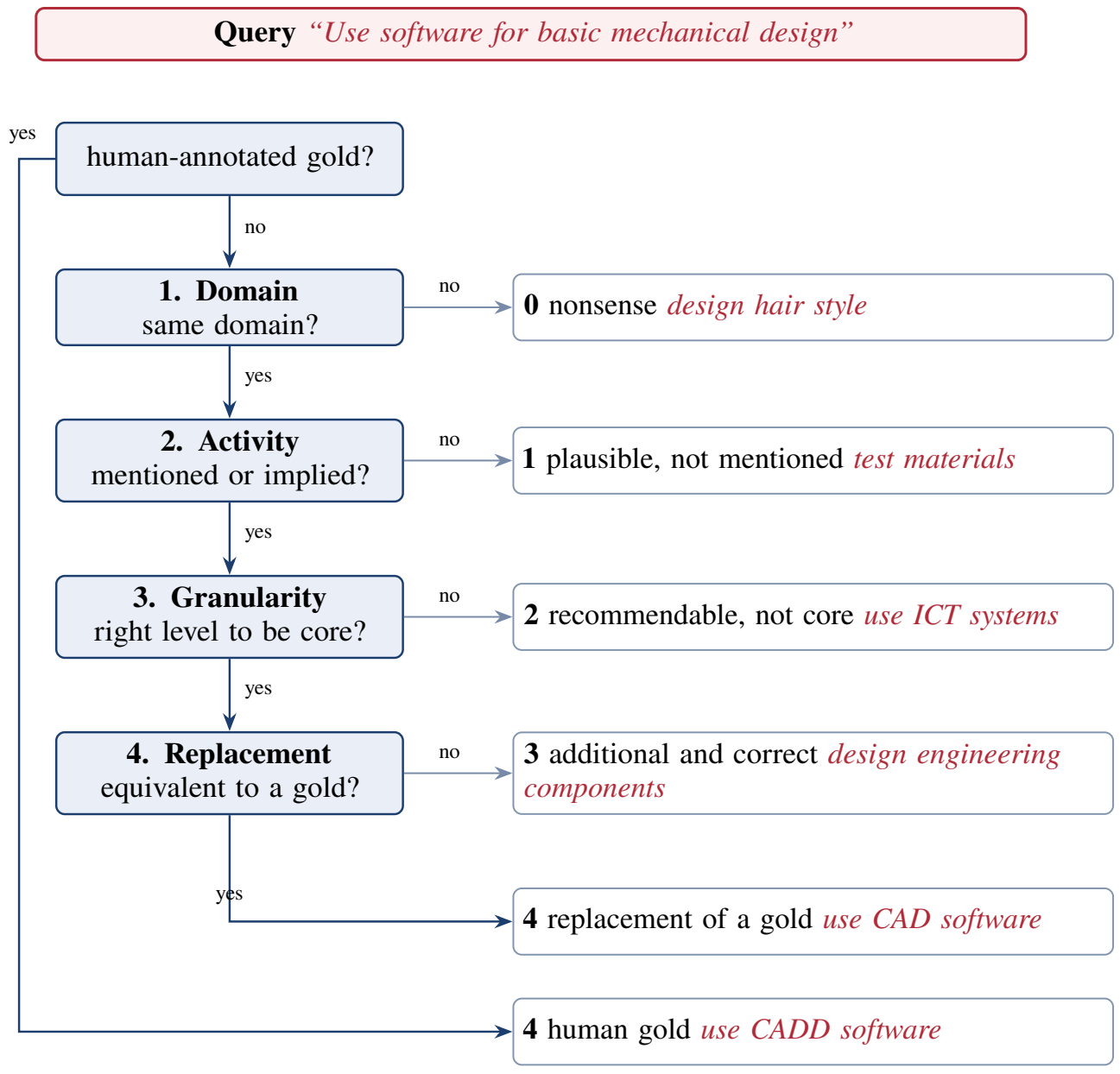
\begin{figure}[t]
\centering
\resizebox{\linewidth}{!}{%
\begin{tikzpicture}[
  font=\scriptsize,
  gate/.style={draw=gateblue,fill=gatefill,rounded corners=2pt,line width=.5pt,
               text width=27mm,align=center,inner sep=2.5pt,minimum height=6mm},
  outc/.style={draw=gateblue!55,fill=white,rounded corners=2pt,line width=.4pt,
               text width=48mm,align=left,inner sep=2.5pt,minimum height=5.5mm},
  ar/.style={-{Stealth[length=1.5mm]},draw=gateblue,line width=.5pt},
  arno/.style={-{Stealth[length=1.5mm]},draw=gateblue!60,line width=.4pt},
]
\hyphenpenalty=10000 \exhyphenpenalty=10000

\node[draw=exred,fill=exfill,rounded corners=2pt,line width=.6pt,text width=80mm,
      align=center,inner sep=3pt,font=\scriptsize] (q) at (0,0)
  {\textbf{Query} \textcolor{exred}{\itshape ``Use software for basic mechanical design''}};

\coordinate (col) at (-25mm,0);
\node[gate,below=5mm of q.south -| col,anchor=north] (g0) {human-annotated gold?};
\node[gate,below=6mm of g0] (g1) {\textbf{1. Domain}\\same domain?};
\node[gate,below=6mm of g1] (g2) {\textbf{2. Activity}\\mentioned or implied?};
\node[gate,below=6mm of g2] (g3) {\textbf{3. Granularity}\\right level to be core?};
\node[gate,below=6mm of g3] (g4) {\textbf{4. Replacement}\\equivalent to a gold?};

\draw[ar] (g0) -- node[right,font=\tiny,pos=.45]{no} (g1);
\foreach \a/\b in {g1/g2,g2/g3,g3/g4}
  \draw[ar] (\a) -- node[right,font=\tiny,pos=.45]{yes} (\b);

\node[outc,right=9mm of g1] (o0) {\textbf{0} nonsense \textcolor{exred}{\itshape design hair style}};
\node[outc,right=9mm of g2] (o1) {\textbf{1} plausible, not mentioned \textcolor{exred}{\itshape test materials}};
\node[outc,right=9mm of g3] (o2) {\textbf{2} recommendable, not core \textcolor{exred}{\itshape use ICT systems}};
\node[outc,right=9mm of g4] (o3) {\textbf{3} additional and correct \textcolor{exred}{\itshape design engineering components}};
\node[outc,below=6mm of g4.south -| o3,anchor=north] (o4) {\textbf{4} replacement of a gold \textcolor{exred}{\itshape use CAD software}};
\node[outc,below=3.5mm of o4.south -| o3,anchor=north] (o5) {\textbf{4} human gold \textcolor{exred}{\itshape use CADD software}};

\foreach \g/\o in {g1/o0,g2/o1,g3/o2,g4/o3}
  \draw[arno] (\g.east) -- node[above,font=\tiny,pos=.42]{no} (\o.west);
\draw[ar] (g4.south) |- node[below,font=\tiny,pos=.28]{yes} (o4.west);
\draw[ar] (g0.west) -- ++(-3mm,0) node[above,font=\tiny,pos=.9]{yes} |- (o5.west);
\end{tikzpicture}%
}
\caption{The annotation cascade, with a worked example on every exit. A pair descends the gates in
order and the first ``no'' fixes its grade, while human gold positives bypass the cascade entirely.
The six red skills are real ESCO entries judged against one real House query, and each is placed by
its ESCO \emph{description} rather than its label. \emph{design hair style} (``design the styles of
hair based on the preference of the customer'') shares only the word ``design'', the kind of surface
overlap a retriever scores highly and the domain gate exists to reject. \emph{test materials} is
engineering work, so it clears the domain gate, but its description covers testing the composition
and characteristics of materials, never designing anything or using software. \emph{use ICT systems}
(``select and use ICT systems for a variety of complex tasks'') is true of the job and of thousands
of others, an umbrella too broad to be core. \emph{design engineering components} (``design engineering parts, assemblies,
products, or systems'') is correct and genuinely additional, describing the design act itself rather
than the drafting tool the gold names. The two grade-4 exits differ only in provenance:
\emph{use CADD software} is the human annotation, and \emph{use CAD software} is the jury judging a
near-paraphrase of it, the two differing only by the drafting suffix.
Section~\ref{sec:filtering} turns that last distinction into the submission screen.}
\label{fig:cascade}
\end{figure}

Direct graded annotation is unreliable: the difference between ``somewhat relevant'' and
``relevant'' is not something two annotators, or two LLM calls, resolve the same way, as
Section~\ref{sec:validation} measures against the same judge and referee. Our rubric therefore asks
four binary questions in a fixed order, the grade being determined by where the cascade stops
(Figure~\ref{fig:cascade}):

\begin{enumerate}
\item \textbf{Domain.} Is the skill in the same broad occupational domain as the sentence? A ``no''
yields grade 0 (nonsense): the pair will likely never appear in the same job description.
\item \textbf{Activity.} Is the specific activity described by the skill actually mentioned or
directly implied? For example, ``theoretical knowledge of mechanical equipment'' does not
imply ``operate mechanical equipment''. A ``no'' yields grade 1 (plausible, but the activity is not
there).
\item \textbf{Granularity.} Is the skill at the right granularity to be core to the sentence,
neither a generic umbrella nor an incidental sub-step? A ``no'' yields grade 2 (could be
recommended, but not core).
\item \textbf{Replacement.} Is the skill semantically equivalent to one of the existing gold
positives? A ``no'' means the skill is genuinely additional and correct: grade 3. A ``yes'' means it
is (close to) a paraphrase of what a human already annotated: grade 4.
\end{enumerate}

Human-annotated gold positives are grade 4 by definition and never reach the judge. Each gate
assumes the previous passed, so every prompt operationalises exactly one decision. The judge always
sees the skill's ESCO description alongside its preferred label, since many labels are ambiguous
alone (``monitor environment'' could be system administration or forestry).

The decomposition buys three things: objectivity, since each question has a defensible yes/no answer
where a direct 0 to 4 grade does not, diagnosability, since a disagreement localises to a single
gate, which is how we iterated on the prompts, and, measurably, accuracy, since on a judge of the
size this workload requires walking the gates beats asking the same model for the grade outright
(Section~\ref{sec:validation}).

\section{Annotation pipeline}\label{sec:pipeline}

\subsection{Candidate generation}

Judging all 13,939 ESCO skills against every sentence is wasteful, since the bulk of the label matrix
is obviously unrelated. We prefilter to the top-2000 candidates per sentence with a BGE-M3
bi-encoder~\cite{bgem3} over sentence-to-description similarity.
Pooled over all datasets and splits this recovers 96.6\% of the human-annotated gold positives
(micro recall@2000). Per dataset: the normalisation dataset is fully covered (100\% on both splits,
trivially so: the prefilter query for these rows is the target concept's own ESCO description, not
the lexically-distant surface form, so the prefilter is never exercised on what makes the dataset
hard), Tech reaches 96.2\% (validation) and 97.6\% (test), House 97.7\% and 93.2\%, SkillSkape
94.7\% and 96.5\%, and TechWolf (test only) 96.4\%. The misses do not concentrate on one failure mode. 
Roughly a third come from long,
multi-topic job-ad paragraphs carrying up to nine gold skills each, where a single pooled
embedding cannot stay close to every skill mentioned. A further quarter fall on generic targets
(\emph{communication}, \emph{work independently}) that collide with near-duplicate ESCO entries.
The remainder are ordinary retrieval misses, some even where a lexical bridge exists: the query
\emph{java reactjs web-services} ranks \emph{Java (computer programming)} only 3{,}681st.  
Skills outside the top-2000 default to grade 0, except should they be a human-annotated gold positive one.
In the latter case the pair is released at grade 4.

\subsection{LLM-as-a-judge cascade}

The four gates are answered by an open-weight LLM judge
(Qwen3-Next-80B-A3B-Instruct~\cite{qwen3,qwen3next}, FP8) served locally on a single Nvidia DGX
Spark. Each gate is one call emitting a single boolean, constrained to a one-field JSON object with
no rationale, so every prompt operationalises exactly one decision and calibration stays gate-local.
%The pipeline sustained roughly 20 gate calls per second end to end, annotating the validation
%release (571K pairs) in under 10 hours and the test release (3.7M pairs) in 68 hours of unattended
%GPU time.

The validation release is single-juror. For the test release we hardened the two most subjective
gates with a second juror from a different model family (Gemma-4-26B-A4B-it~\cite{gemma4}),
re-judging gate 3 and, for its survivors, gate 4, on only the pairs the first juror passed through
gate 2. A gate passes only if both jurors agree, so the second can demote a grade 3 or 4 but never
promote it, firming up the soft upper gates without touching the high-volume lower ones. This costs
46K extra calls (about 1\% of first-pass volume) and demoted 6,449 of the 31,557 pairs it
re-examined. The test cascade also ships a recalibrated gate-3 prompt, though
Section~\ref{sec:validation} shows the resulting labels behave near-identically to the validation
configuration under blind review.

\subsection{Prompt calibration}\label{sec:calibration}

Prompts were calibrated iteratively against hand-graded pilot pairs. After a first full pass over
the validation annotations, we used Claude Opus 4.7 to (1) cluster the most frequent failure cases
per gate and (2) interactively probe the local judge to tune the corresponding prompt. After the
validation release, a human iteration over the same judge-of-the-judge loop refined the prompts once
more before the test set was annotated.

Three recurring failure modes drove the concrete prompt changes. At the activity gate, the judge
conflated qualified and unqualified activities, failing ``manage ICT project'' against a sentence
mentioning generic ``project management'' even though the mismatch is one of granularity: the fix
strips sector, method, audience and instance qualifiers to a bare verb-noun pair before checking
presence, deferring the qualifier judgement to gate 3 where it belongs. The same gate needed a
carve-out for job-description phrasings, where ``experience with X'' and ``proficient in X'' license
the activity X. At the granularity gate, the validation-era prompt defaulted too far toward
rejection, treating any skill phrased more generally than the sentence as too broad. The test-era
recalibration reframes it as ``a filter, not a hurdle'' while hardening the opposite direction with
an unnamed-specific-instance rule: a generic category mention (``a modern programming language'')
never licenses one specific member (Python). Finally, the released cascades drop the one-sentence
rationale an early variant asked for, after a pilot showed the boolean-only variant stays within one
grade on 95.6\% of pairs, the residual disagreements concentrating on 2 to 4 word ``section header''
inputs where human annotators also disagree.

\subsection{Human validation of the labels}\label{sec:validation}

To validate the released labels, a fixed, stratified manifest of pairs (280 test, 224 validation,
per dataset 12 pairs of each grade 1 to 4 and 8 of grade 0) was blindly re-graded under the
identical rubric: no cascade label shown, one independent judgement per pair. There are two
reviewers: a frontier model from a different family than the cascade judge (Claude
Opus 4.8~\cite{claudeopus}), covering both splits, and a human expert, covering the 280-pair test
manifest.

Grades 3 and 4 are reported together throughout this section. The only thing separating them is
gate 4, which asks whether the skill is a replacement for one the dataset already carries as a human
gold positive: that is a property of the pre-existing annotation rather than of the (sentence, skill)
pair, and a blind reviewer cannot see the gold list, so the distinction is not one they are
positioned to confirm. Collapsing it leaves the three substantive gates (domain, activity,
specificity) as what the validation measures. Gate 4 is not noise: the pairs it promotes draw a
higher mean blind grade than the grade-3 pairs it rejects (2.63 versus 2.22 from the human expert,
2.48 versus 2.20 from Opus). It is simply not a distinction this validation can referee.

\begin{figure*}[tb]
\centering
\begin{minipage}[t]{0.32\linewidth}
\centering
\includegraphics[width=\linewidth]{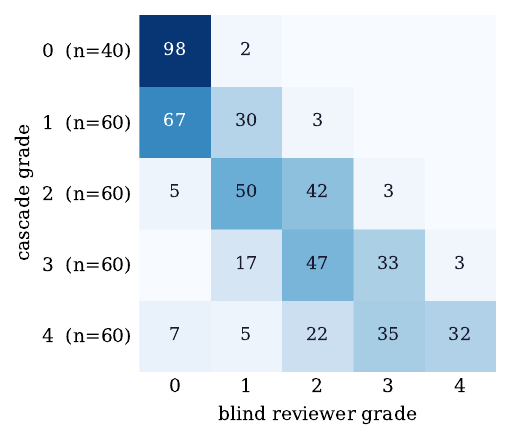}
\end{minipage}\hfill
\begin{minipage}[t]{0.32\linewidth}
\centering
\includegraphics[width=\linewidth]{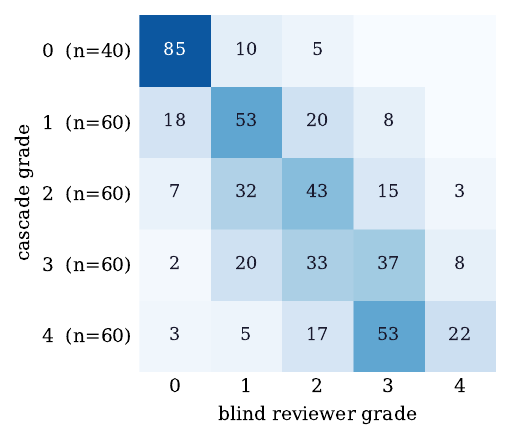}
\end{minipage}\hfill
\begin{minipage}[t]{0.32\linewidth}
\centering
\includegraphics[width=\linewidth]{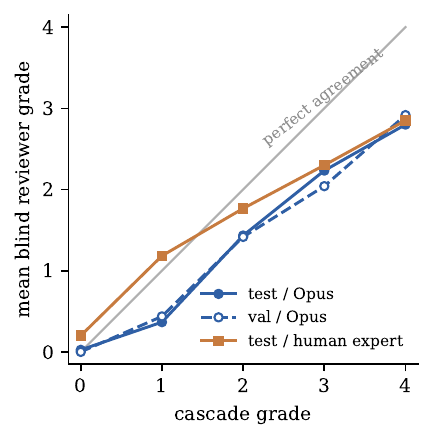}
\end{minipage}
\caption{Blind validation of the cascade labels, test manifest. \emph{Left}: row-normalised
confusion matrix (\%) between the released cascade grade and the blind Opus reviewer. \emph{Middle}:
the same, against the blind human expert. Both show mass on the diagonal or one cell to its left
with a nearly empty upper-right triangle: the disagreement is one-directional for both reviewers,
human included. \emph{Right}: mean blind grade per cascade grade, Opus (both splits) and the human
expert (test only). All three curves sit below the identity line, and the human curve is markedly less
extreme at grade 1 but has the same overall shape. Grades 3 and 4 are reported together throughout,
since grade 4 is defined against the gold-positive list the blind reviewer cannot see.}
\label{fig:validation}
\end{figure*}

Four findings (Figure~\ref{fig:validation}). First, ordering is strongly preserved between the
cascade and both reviewers. Quadratic-weighted kappa (QWK), which is chance-corrected agreement that
charges a disagreement in proportion to its squared distance on the grade scale, is $0.73$ (Opus)
and $0.69$ (human); Spearman rho, the rank correlation between the two gradings, is $0.82$ and
$0.69$; 93\% and 90\% of pairs are within one grade and exact agreement is 51\% and 59\%. Plain
Cohen's kappa ($0.36$ and $0.43$) is the wrong lens for an ordinal scale, punishing a 2-against-3
miss as harshly as a 0-against-3 one. Second, the disagreement is monotonic for both
reviewers: the cascade is generous 47\% of the time against Opus and 29\% against the human, against
only 2\% and 12\% too strict, and grade-0 negatives are highly reliable for both (98\% and 85\%
confirmed). Third, the bias is stable across cascade versions, reproducing near-identically on both
splits (kappa $0.73$ test against $0.68$ validation). Fourth, and this is what closes the
validation, the human expert, sharing neither the reviewer's model family nor its rubric prompt,
reproduces the same one-directional generosity, so it is a property of the LLM-cascade approach
rather than an artifact of one model grading another. Comparing human and Opus directly is
consistent: the human grades higher on the same pair 34\% of the time against the reverse 11\%,
sitting between Opus and the cascade rather than in an independent third direction. The one genuine
disagreement, rather than added noise, is at grade 1 (``weak''), graded far less harshly by
the human (mean blind grade $1.18$ against Opus's $0.37$).

What the labels are asked to do is order candidates within a query, not to fix an absolute grade,
so an offset that is monotonic and the same for every system leaves that ordering intact and some
noise in the exact grade is acceptable. We therefore document the offset rather than re-label. The
challenge metric is graded nDCG, and for any binary relevant/irrelevant cut we recommend grade
$\geq 2$, where binary agreement with the blind reviewers peaks.

The cascade spends up to four calls per pair to reach a grade a single call could be asked for
directly, so it is fair to ask what the decomposition buys. We ran the \emph{same} juror over the
same 280-pair test manifest in one call, emitting the whole grade from a prompt that carries the
calibrated content of gates 1 to 3 and states the decision order explicitly, and scored both arms
against the blind human expert. Everything outside the decomposition is held fixed: same weights,
same greedy decoding, same structured output, byte-identical user prompt. Neither arm sees the
gold-positive list, so both are judged on the 0 to 3 scale used above.

The cascade wins on every agreement measure (Table~\ref{tab:ablation}). Against the rationale
variant it is $+0.068$ QWK and $-0.125$ in mean absolute error (MAE, the average distance in grades
between judge and human), with 95\% confidence intervals of $[+0.006, +0.132]$ and
$[-0.204, -0.046]$ from a paired bootstrap over pairs (10{,}000 resamples, stratified by cascade
grade), and a signed-rank test on the per-pair absolute errors at $p = 0.005$. Against the
grade-only variant the MAE gap holds at $p = 0.002$ while the QWK gap is directionally the same but
not significant. The one-shot arms also use the middle of the scale less consistently, and which
intermediate grade thins out moves with the decoding format, where a cascade grade is a consequence
of where a pair stops. The claim we draw is the narrow one: at the model scale this annotation
workload requires, decomposing the grade into ordered binary gates produces labels closer to a human
expert than asking the same model for the grade directly.

\begin{table}[t]
\centering
\small
\setlength{\tabcolsep}{3.5pt}
\caption{Cascade versus one-shot direct grading by the same judge, against the blind human expert on
the 280-pair test manifest (grades 0 to 3). Gen./Str. is the share of pairs the judge grades above or
below the human.}
\label{tab:ablation}
\begin{tabular}{l c c c c c}
\toprule
\textbf{Judge} & \textbf{QWK} & \textbf{$\rho$} & \textbf{Exact} & \textbf{MAE} & \textbf{Gen./Str.} \\
\midrule
Cascade (released)   & \textbf{0.686} & \textbf{0.686} & \textbf{0.586} & \textbf{0.529} & 29/12 \\
One call, rationale  & 0.618 & 0.621 & 0.525 & 0.654 & 25/22 \\
One call, grade only & 0.631 & 0.627 & 0.461 & 0.675 & 22/32 \\
\bottomrule
\end{tabular}
\end{table}

\subsection{An ESCO caveat}

Some ESCO preferred labels do not carry a restriction that their description imposes, so a pair can
look like a strong match from the label alone and be graded down once the description is read.
Because the judge always sees both, the released grades track the described concept rather than the
label string. One instance is \emph{guide staff}, a label that reads as generic people management.
Its description is ``Lead and manage team in order to keep them informed about a variety of rules
and regulations regarding to grants'', which ties it to grant administration. Over the 1{,}873
pairs it was judged on it never receives a grade above 2, and the 33 pairs that reach 2 are
team-leadership queries spread over all four extraction datasets, among them \emph{Can you motivate
and set the direction for your employees?} and \emph{strong people management or team leadership
skills}. That ceiling is only defensible from the description. On the first of those queries the
rest of the grade-2 tier are labels that state their restriction openly, such as \emph{supervise
dental staff}, \emph{lead a team in forestry services} and \emph{manage zoo staff}, while the
labels graded 3 and 4 are unrestricted ones such as \emph{lead others} and \emph{supervise staff}.
Read as a string, \emph{guide staff} belongs with the latter. The same holds for \emph{apply conflict
management}, graded 2 against \emph{outstanding negotiation skills with the ability to resolve
issues and address complaints}, because its description is about handling ``a problematic gambling
situation''. Anyone inspecting or training on the released labels should read them alongside the
descriptions.

\section{Skill extraction datasets}\label{sec:extraction}

\begin{table}[t]
\centering
\footnotesize
\setlength{\tabcolsep}{3pt}
\caption{The five released benchmarks. Queries are ranked against the same corpus of $13{,}891$ ESCO
v1.1.0 skills, so the released qrels are the full query $\times$ corpus grid, with every pair the
prefilter did not surface carried at grade 0. Judged pairs counts the pairs the cascade actually
graded (queries $\times$ the top-2000 window, $4.3$M in total). The last two columns are test-split
means per query: targets graded $\geq 2$, and targets graded 4. TechWolf ships no validation
split.}
\label{tab:corpus}
\begin{tabular}{l r r r r r}
\toprule
 & \multicolumn{2}{c}{\textbf{Queries}} & \textbf{Judged} & \multicolumn{2}{c}{\textbf{Test, per query}} \\
\cmidrule(lr){2-3}\cmidrule(lr){5-6}
\textbf{Dataset} & \textbf{Val} & \textbf{Test} & \textbf{pairs} & \textbf{gr.\ $\geq 2$} & \textbf{gr.\ 4} \\
\midrule
House      &  61 & 261 & 0.64M & 22.7 & 2.7 \\
Tech       &  75 & 338 & 0.83M & 15.8 & 1.9 \\
TechWolf   & n/a & 324 & 0.65M & 17.5 & 2.3 \\
SkillSkape & 100 & 500 & 1.20M & 27.6 & 3.2 \\
SkillNorm  &  50 & 450 & 1.00M &  9.3 & 1.1 \\
\midrule
Total      & 286 & 1,873 & 4.32M & & \\
\bottomrule
\end{tabular}
\end{table}

\subsection{Dataset selection}
%%%% till here I've read it
Four of the five datasets are established ESCO skill extraction benchmarks, with vacancy sentences
as queries, re-annotated with graded relevance. Table~\ref{tab:corpus} summarises the five released
benchmarks.
House and Tech~\cite{decorte2022negative} are annotated exhaustively on both validation and test
splits, as is the TechWolf test set~\cite{decorte2025context} (which ships no validation split).
SkillSkape~\cite{skillskape} is subsampled to 100 validation and 500 test sentences (from 2,751 and
3,244 source rows) because exhaustive graded annotation of the full splits is prohibitive: at
roughly 2,000 judged candidates per sentence, the full SkillSkape validation split alone accounts
for an order of magnitude more judge calls than the other datasets combined.

\paragraph{Subsampling SkillSkape.} The publicly released SkillSkape splits carry $2{,}751$ and
$3{,}244$ (sentence, label) rows over $1{,}316$ and $1{,}272$ unique sentences, spread across $149$
and $455$ distinct ESCO labels, so the median label occurs only $14$ and $5$ times and the test tail
reaches single occurrences. Keeping just $7.6\%$ of validation and $39\%$ of test sentences, naive
sentence-level sampling would drop rare labels entirely and distort the labels-per-sentence
distribution. We therefore bucket sentences by label cardinality ($\{1,2,3,4,5{+}\}$), give each
bucket a quota proportional to its share of the source split with largest-remainder rounding, and
fill each bucket by iterative stratification~\citep{sechidis2011stratification}, repeatedly taking
the globally rarest still-needed label and picking a sentence that carries it while also helping
fill quotas for its other labels (ties broken by largest residual quota, then a fixed seed). The
result preserves the label marginals (L1 distance $0.58$ and $0.41$ to the source distributions) and
the labels-per-sentence means ($2.10$ and $2.56$ against $2.09$ and $2.55$), while retaining $77\%$
and $94\%$ of the two label spaces. We keep the original long format, so the samples are drop-in
compatible with the SkillSkape loader.

\section{Skill normalisation dataset}\label{sec:norm}

The fifth dataset targets skill normalisation. Its queries are not vacancy sentences but surface
forms of individual ESCO skills, and the task is unchanged: the full vocabulary is ranked, every
candidate in the retrieval window carries a graded label from the same cascade, and the skill's
preferred label is the single gold positive. To increase the difficulty, we sample surface forms
that share the fewest words with their ESCO preferred label. ESCO's preferred labels each come with a
set of alternative labels, but most
are uninteresting morphological or syntactic variants (e.g.,
\emph{monitor crops} vs.\ \emph{monitoring crops}) that test little beyond
tokenization.  To construct a more challenging skill normalisation benchmark, we
sample alternative labels whose surface form diverges substantially from the
preferred label while still denoting the same concept.

\paragraph{Two filtering gates.} We first score each (preferred, alternative) pair with a
\emph{stem-Jaccard divergence}: both strings are lowercased, tokenized, stripped of stopwords,
normalized for British/American spelling, and reduced to stems by iterative suffix stripping, with
the divergence taken as $1 - |A \cap B| / |A \cup B|$ over the stem sets. We retain pairs with
divergence $\geq 0.7$ that also have at least two content tokens on each side, a token-count ratio
in $[0.4, 2.5]$, and no all-caps abbreviations. Of ESCO v1.1.0's $85{,}270$ alternative labels,
$14{,}423$ pairs over $5{,}232$ preferred labels survive.

A purely lexical filter is insufficient: manually reviewing $30$ pairs from this pool showed roughly
$10\%$ are not valid paraphrases at all, either because ESCO itself lists a related but distinct
concept as an alternative (\emph{attend to fitness clients under controlled health conditions}
alongside \emph{be aware of professional standards and limitations}) or through incidental
near-misses (\emph{assess seizable goods} against \emph{appraise sizable goods}, a typo turned
synonym). Stem-Jaccard cannot catch either, since both have low surface overlap by construction. We
therefore add a semantic gate, embedding each pair with
\texttt{all-mpnet-base-v2}~\cite{sentencebert,mpnet} and requiring cosine $\geq 0.5$. The two gates
are complementary: divergence makes a pair benchmark-\emph{interesting}, cosine makes it
benchmark-\emph{valid}. The cosine gate removes $20.7\%$ of the lexically eligible pool, leaving
$11{,}434$ pairs over $4{,}632$ preferred labels, from which we sample $500$ uniformly at random with
at most one alternative per preferred label (50 validation, 450 test), yielding mean divergence
$0.84$ and mean cosine $0.72$. Table~\ref{tab:altlabel-scores} shows the two scores acting in
concert.

\begin{table}[t]
\centering
\footnotesize
\setlength{\tabcolsep}{3pt}
\caption{ESCO preferred/alternative pairs at each filtering stage. Divergence $\geq 0.7$ removes
morphological variants (middle block), and cosine $\geq 0.5$ removes pairs sharing no semantic content
despite different wording (bottom block). The top block satisfies both.}
\label{tab:altlabel-scores}
\resizebox{\linewidth}{!}{%
\begin{tabular}{l l c c}
\toprule
\textbf{Preferred label} & \textbf{Alternative label} & \textbf{Div.} & \textbf{Cos.} \\
\midrule
\multicolumn{4}{l}{\emph{Retained}}\\
provide periodontal treatment & manage gum conditions & 1.00 & 0.64 \\
install crane equipment & build lifting machinery & 1.00 & 0.66 \\
pest management techniques & extermination methods & 1.00 & 0.56 \\
monitor crops & checking crop growth & 0.75 & 0.74 \\
welcome restaurant guests & greet and seat guests & 0.83 & 0.68 \\
\midrule
\multicolumn{4}{l}{\emph{Rejected: low divergence}}\\
perform toxicological studies & carry out toxicological studies & 0.50 & n/a \\
monitor crops & crop monitoring & 0.00 & n/a \\
handle equipment suspended & handling equipment suspended & 0.00 & n/a \\
\midrule
\multicolumn{4}{l}{\emph{Rejected: low cosine}}\\
attend to fitness clients (\ldots) & be aware of standards (\ldots) & 1.00 & 0.27 \\
tend mechanical forging press & guard machined fashioned stamp & 1.00 & 0.23 \\
assess seizable goods & appraise sizable goods & 0.80 & 0.42 \\
\bottomrule
\end{tabular}}
\end{table}

Grading for this dataset runs against the \emph{canonical concept} rather than the surface
form: both the prefilter query and the sentence shown to the judge are built
from the preferred label and its ESCO description, and the alternative label is
withheld. Participants are served the alternative label alone as the query.

\section{Challenge setup}\label{sec:setup}

The challenge runs on CodaBench\footnote{\url{https://www.codabench.org/competitions/16900/}} with an
open validation phase (unlimited submissions) followed by a test phase capped at 10 submissions per
team, and drew 63 registered participants. The leaderboard metric is nDCG@100 against the graded
labels, averaged over the five datasets. All evaluation goes through
WorkRB~\cite{delange2026workrb}: a participant model implements the library's
\texttt{ModelInterface}, a single ranking method over queries and targets, and submission files come
from the same WorkRB tasks the organisers use. The rules are deliberately short: any technique
producing rankings competes, training is optional, any open-source data except test set information
is allowed, and the one expectation is a contribution of the team's own rather than an unmodified
off-the-shelf model.

%The starter kit ships a trainable bi-encoder baseline
%(paraphrase-mpnet-base-v2~\cite{sentencebert,mpnet} with in-batch InfoNCE on 138K synthetic
%sentence-skill pairs), worked examples of a cross-encoder, a frozen-encoder classifier head and an
%inference-only retrieve-then-rerank pipeline, plus parameter-free lexical baselines (TF-IDF,
%BM25~\cite{bm25}) participants can score locally. Because the metric is nDCG@100, re-ordering the
%first hundred candidates is where the points are, which makes rerankers a cheap entry route.

\paragraph{Availability.} All five graded benchmarks, validation and test splits alike, are openly
released under CC BY 4.0 on the Hugging Face Hub\footnote{\url{https://huggingface.co/TechWolf}}
and are served through the WorkRB package as the same tasks the leaderboard scored.
Each release carries the full query $\times$ corpus grid of graded labels rather than the top-2000 window alone,
so the test labels this paper analyses are reusable as a benchmark after the challenge.

\paragraph{On the baselines.} The two trained baselines above (ConTeXT-match and the shipped
CurriculumMatch retriever) enter the leaderboard as released checkpoints, so a team that beats one
has beaten a particular artefact rather than the method. We checked how much of the gap is
deployment: retraining ConTeXT-match from the authors' code reproduces the checkpoint ($0.610$
against $0.607$), and holding method, data and loss fixed while tuning only the training schedule
and selecting the checkpoint on the graded validation split reaches $0.620 \pm 0.003$ over three
seeds.

\section{Submission filtering}\label{sec:filtering}

The five test benchmarks also exist publicly with their original \emph{binary} annotations, which
the rules put off-limits at every stage. We therefore screen the submissions before reporting any
result. 

The screen exploits a property of the cascade: Grade 4 is a union of two
populations: a pair is graded 4 either because it \emph{is} a gold positive of the underlying
dataset, which never reaches the jury (4-human), or because the jury judged it a direct replacement
of one at the final gate (4-jury). The two are semantically (almost) equivalent by construction, and only the
first was public during the challenge. For each query we compare the ranks a submission gives its 4-human items
against those it gives its 4-jury items, summarised as the probability that a random 4-human item
outranks a random 4-jury one, ties counting one half. This normalised Mann-Whitney statistic has an
expected value of $0.5$ for any system with no access to the public files.

The statistic is sharply bimodal on all four extraction datasets with an empty band between the
modes. The lower mode sits at the null (medians $0.60$, $0.59$, $0.63$ and $0.48$ on Tech, House,
TechWolf and SkillSkape, the mild excess being the cascade's known generosity on positives). The upper
mode is not a matter of degree: 57 submissions attain a separation of exactly $1.0$ on all
four datasets simultaneously, meaning every public positive outranks every non-public grade-4 item
on every query. The same 57 also place the public positives in the top $k$ of every query, $k$ being
that query's number of public positives, so their ranking reproduces the withheld annotation exactly.
Three properties argue the statistic reads provenance: within the lower mode it correlates
\emph{negatively} with nDCG@100 (Spearman $-0.40$), so competence does not raise it and
the strongest unflagged system reaches only $0.56$, the four datasets, with disjoint queries and
independent annotation runs, agree on whom to flag at Jaccard $0.967$ to $1.0$, and twelve affected
teams have both flagged and unflagged entries, so it is not a property of a team's ability. We
exclude SkillNorm from the verdict, where roughly one gold concept per query saturates the statistic
near $1.0$ for everyone, clean teams included.

Of the 43 teams with at least one scored test submission (218 in total), the screen removes
61 submissions ($28.0\%$) from 15 teams. Because most affected teams also submitted
unflagged entries, only 3 teams lose all of theirs and 40 of 43 remain in the
reported results. The effect on the ordering is larger than the counts suggest, as the removed
entries concentrate at the top. Section~\ref{sec:results}
reports the filtered board, each team represented by its best remaining submission. The screen leaves
157 clean submissions; setting aside the two organiser baselines, which are reported as their own rows,
and one entry uploaded twice, this leaves 154 Test-phase submissions from 39 participant accounts.

\section{Results and insights}\label{sec:results}
\subsection{Leaderboard}

\begin{table*}[t]
\centering
\small
\setlength{\tabcolsep}{2.5pt}
\caption{Filtered leaderboard: graded nDCG@100 per task and macro-averaged, best valid submission
per account, plus the organiser baselines. Teams are named by their CodaBench handle throughout. 
The winning entry is the work of Maksim Silchenko (National University of Singapore), 
competing solo as team \texttt{thylinao},
who documents the system in~\cite{silchenko2026ranking}. $^{s}$~self-declared student team. The ConTeXT-match
row is the pre-released checkpoint. Retrained with a tuned schedule the
same method reaches $0.620$ (Section~\ref{sec:setup}).}
\label{tab:leaderboard}
\begin{tabular}{r l c c c c c c}
\toprule
\textbf{\#} & \textbf{Team} & \textbf{Macro} & \textbf{House} & \textbf{Tech} & \textbf{TechWolf} & \textbf{SkillSkape} & \textbf{SkillNorm} \\
\midrule
1 & thylinao$^{s}$      & \textbf{0.8458} & 0.8387 & 0.8390 & 0.8347 & 0.8683 & 0.8480 \\
2 & qihang-wang         & 0.8323 & 0.8405 & 0.8337 & 0.8197 & 0.8585 & 0.8092 \\
3 & yeyang$^{s}$        & 0.8224 & 0.8226 & 0.8261 & 0.8030 & 0.8586 & 0.8018 \\
4 & aiden-lee           & 0.8191 & 0.8114 & 0.7903 & 0.8109 & 0.8316 & 0.8514 \\
5 & elara96             & 0.8182 & 0.8015 & 0.8157 & 0.8022 & 0.8443 & 0.8274 \\
6 & carolzz             & 0.8132 & 0.7982 & 0.8106 & 0.8026 & 0.8442 & 0.8101 \\
7 & liyitang$^{s}$      & 0.8131 & 0.8014 & 0.8155 & 0.8025 & 0.8315 & 0.8144 \\
8 & jint1234$^{s}$      & 0.8119 & 0.8172 & 0.7866 & 0.7958 & 0.8454 & 0.8145 \\
9 & mingc               & 0.8116 & 0.7907 & 0.7908 & 0.7954 & 0.8302 & 0.8508 \\
10 & junlong$^{s}$      & 0.8112 & 0.8170 & 0.8093 & 0.8152 & 0.8444 & 0.7701 \\
\midrule
n/a & CurriculumMatch & 0.6292 & 0.5746 & 0.6022 & 0.6056 & 0.7014 & 0.6620 \\
n/a & ConTeXT-match   & 0.6075 & 0.5754 & 0.6020 & 0.5918 & 0.6337 & 0.6345 \\
n/a & TF-IDF & 0.2679 & 0.2503 & 0.3053 & 0.2614 & 0.2499 & 0.2727 \\
n/a & BM25            & 0.2250 & 0.2042 & 0.2773 & 0.2124 & 0.1724 & 0.2589 \\
\bottomrule
\end{tabular}
\end{table*}

Table~\ref{tab:leaderboard} shows the filtered leaderboard. The top is dense: twelve accounts finish
within $0.04$ of the winner and positions 6 to 11 span $0.003$. Beating the trained baselines was
routine (32 of 39 accounts clear ConTeXT-match, 31 clear CurriculumMatch), but the margin above them
is real: the winner's $0.8458$ sits $0.217$ above the strongest baseline, and
Section~\ref{sec:size} prices what the field spent to buy it. The two lexical baselines bracket the
floor, with two submissions being BM25 reimplementations that land within $0.002$ of it.

\subsection{What the graded labels reveal}\label{sec:graded-vs-binary}

\begin{figure}[t]
\centering
\includegraphics[width=\linewidth]{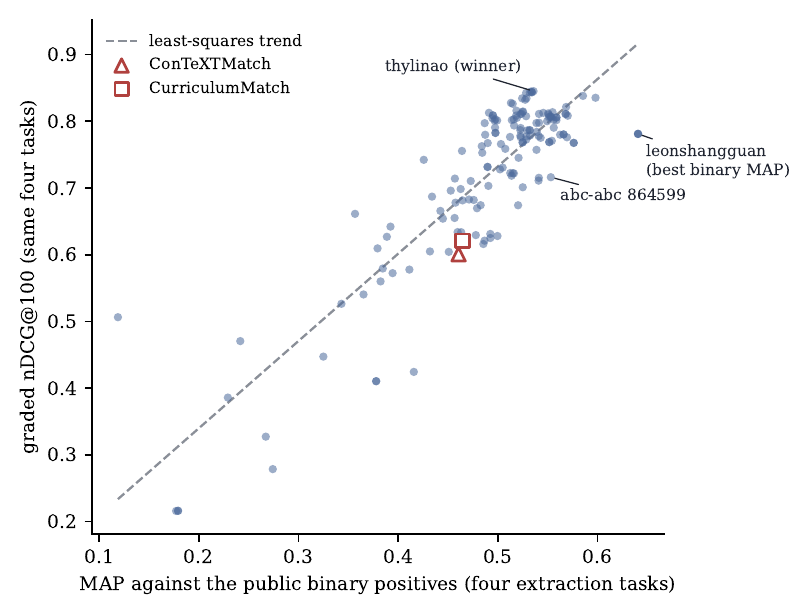}
\caption{Graded nDCG@100 against MAP on the public binary positives, both macro-averaged over the
four extraction tasks (SkillNorm has no binary counterpart and is excluded from both axes). One
point per valid submission. The two metrics agree on the broad strokes and disagree exactly where
a leaderboard matters: the graded winner is 12th by binary MAP, and the binary-MAP leader
(leonshangguan) is 16th on the graded board.}
\label{fig:graded-vs-binary}
\end{figure}

\begin{figure}[t]
\centering
\includegraphics[width=\linewidth]{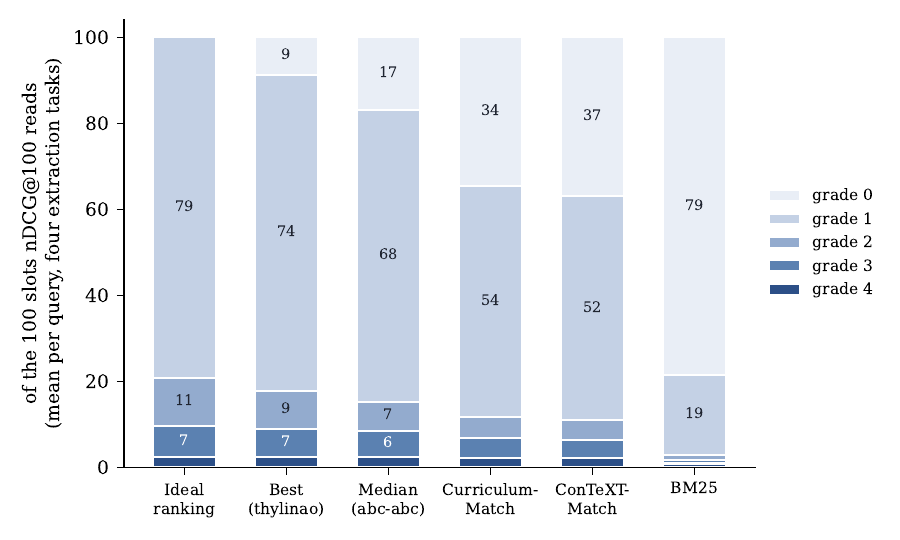}
\caption{Grade composition of the 100 slots that nDCG@100 reads (mean per query, macro over the
four extraction tasks) for the ideal ranking, the best and median valid submissions, and the
organiser baselines. The ideal window contains essentially no grade-0 items, and the ranking systems
differ mostly in how many grade-0 items they let in (9 for the winner, 17 at the median, 34--37
for the trained baselines, 79 for BM25) while everyone finds nearly all $\approx$2.5 grade-4
slots.}
\label{fig:grade-profile}
\end{figure}

The core question the annotation effort exists to answer: does grading the whole window change the
verdict binary evaluation would have given? For each valid submission we compute MAP against the
original public binary positives of the four extraction datasets alongside graded nDCG@100 on the
same four tasks (Figure~\ref{fig:graded-vs-binary}).\footnote{Submissions consist of only their top slice (mostly 100), and
positions below it are completed in ConTeXT-match order. Submissions themselves rank a median of
96\% of the binary positives, and zero-filling instead moves no submission's MAP by more than
$0.0023$.} Field-wide the two agree (Spearman $\rho = 0.72$ over 154 submissions, $0.80$ over the 39
team bests), which is expected of two metrics that both reward finding relevant skills, and is not
the interesting result.

The agreement dissolves exactly where a leaderboard needs it. Over the twelve best teams the two
orderings are uncorrelated: Kendall $\tau = 0.06$ (top twenty: $0.02$). Under binary MAP the winner
would be leonshangguan ($0.641$), whom the graded metric places 16th, while thylinao, the graded
winner, is 12th by MAP. The binary protocol also compresses the gap to the baselines: the winner
clears the ConTeXT-match checkpoint by $0.244$ graded nDCG@100 (41\% relative) but only $0.075$ MAP
(16\% relative). 

The mechanism is the one \citet{veys2026surface} predicted: MAP asks only where the annotated
positive landed, and nDCG@100 charges for the other 99 slots. The starkest pair makes it concrete. A
mid-table submission (\texttt{team-1}, $0.7143$ macro) \emph{beats the winner on binary MAP and MRR}
($0.553$ against $0.536$, and $0.666$ against $0.652$) while losing $0.131$ graded nDCG. The two
metrics read disjoint parts of the same window: MAP is settled by the single annotated positive,
which both systems place at essentially the same rank, so it is near-blind to the other 99 slots,
whereas nDCG@100 scores every one of them and so records what the two systems do differently.
Concretely, the challenger places an explicitly graded-0 target in its top
five on 15\% of queries against the winner's 4\%. Its errors are lexical-shortcut errors of exactly
the kind the grading was built to expose: \emph{landscape analysis} for a sentence using
``landscape'' figuratively, \emph{operate cutting equipment} for a ``cut-over'' deployment,
\emph{computational fluid dynamics} for ``microsoft-dynamics''. 

The winner's own ablations price what the binary protocol costs a participant who trains against
it. Fine-tuning a MiniLM-L6 cross-encoder on binary targets measured $-0.0758$ graded nDCG@100
\emph{against the base it was meant to improve}~\cite{silchenko2026ranking}: binary supervision
drives grades 1 and 2 to a hard zero, discarding the partial credit those grades still pay under
exponential gain, and the ordering that remains is worse than the one the untuned model already
had. The winner's own audit of the failure names the mechanism: in the retrieval rank window it had
mined negatives from, 60--72\% of the items it was labelling 0 were in fact graded gold, so the loss
was explicitly trained to bury targets that earn credit. A second team measures the same sign
independently, a generic binary cross-encoder taking its macro from $0.649$ to $0.611$, and reaches
the same diagnosis: a binary target flattens the $1 < 2 < 3 < 4$ ordering the metric pays for. This
is the failure \citet{veys2026surface} diagnosed, measured on the metric that charges
for it: there, mitigating the shortcut \emph{lowered} MAP ($44.9 \to 43.1$) while cleaning the
window (judged-nonsense rate $11.5 \to 9.5$), so the binary metric penalised the better ranker.

Figure~\ref{fig:grade-profile} shows where the points are won and lost. Every system, lexical
included, recovers nearly all of the $\approx$2.5 grade-4 slots the ideal top-100 contains:
ranking the human positives highly is the part of the task binary training data already teaches,
and BM25 still recovers 36\% of those slots against only $\sim$10\% of the grade-2/3 ones,
because gold positives are heavily enriched in surface overlap: on Tech, 13.9\% of grade-4
targets have their whole label spelled out verbatim in the sentence, against 0.03\% of grade-1
targets. What
separates the field is the tail of the window: the share of top-100 slots spent on explicitly
irrelevant targets runs 9 (winner), 17 (median), 34--37 (trained baselines), 79 (BM25) out of
100. The graded metric's extra discriminative power over the binary protocol is almost entirely
this quantity: it measures how much nonsense a system tolerates once the easy positives are
placed.

\subsection{Method families}\label{sec:methods}

The field converged on one architecture. Two sources document it: source code, which 13 accounts
shipped and we read in full (11 of them survive the screen of Section~\ref{sec:filtering}), and the
free-text system description CodaBench required with every submission, which for a further 22
reported accounts names the components explicitly. Together these document 33 of the 39 reported
accounts and 19 of the top 20; the rest describe their entries only as version tags. Both sources
tell the same story. Every code-shipping account runs a multi-stage retrieve-then-rerank pipeline:
a fleet of dense retrieval lanes, fused, then re-ordered by cross-encoder or LLM rerankers, with a
learned fusion layer (LambdaRank-family gradient boosting in at least five of them) on top, and
every one of them, both prize positions included, carries lexical retrieval or lexical features
alongside its dense lanes. Of the 22 description-only accounts, 16 name a reranking stage over a
retrieved shortlist and 4 more a multi-lane retrieval ensemble without one, so 27 of the 33
documented systems are retrieve-then-rerank and none departs from the retrieve-and-fuse skeleton.
The winner is representative: twelve retrieval lanes and four fine-tuned cross-encoder columns
fused by a LightGBM LambdaRank bag. The variation is in the reranking stage, and it spans three
families: fine-tuned cross-encoder columns (thylinao, qihang-wang: a Qwen3-8B and BGE cascade
trained on self-distilled labels), hosted-LLM rerankers (pjmathematician reranks its top-200 with
Claude, training-free, at $0.8109$, and leonshangguan spends $\sim$11K frontier-model calls for a
measured $+0.049$), and graded-objective rerankers trained on the released validation labels.
Purely data-centric or single-encoder entries populate the bottom half of the board.

\paragraph{Trainings data.} Any other open-source data was allowed and the five test benchmarks in any form
were not. The field used a narrower slice than that. Across the 13 code submissions, 11 train on
the graded validation labels, 9 on the TechWolf synthetic ESCO skill sentences, and only 5 reach for any public corpus
beyond those two. The 286 validation queries are the contested resource: two systems
(\texttt{aiyuanwang}, \texttt{ccmike}) declare them their \emph{only} supervision, \texttt{ccmike}
enforcing that in code with a data-access policy that refuses to open anything else, while the two
teams declining them, including \texttt{carolzz} in 6th, both say they wanted one honest evaluator
left although the rules permit their use. Where more data was wanted the field manufactured it: five systems build graded supervision by having a larger model grade
candidate pairs, one re-grading only the zero-labelled rows with a three-model jury, another
pseudo-grading all 138K synthetic queries with ordinal-calibrated cutoffs. The winner tried the same
and reports it failing its acceptance gate at $-0.0094$, a content mismatch rather than a quality
problem: its synthetic House queries came out easier than the real ones, so the rows did not sit on
the $+0.013$-per-doubling curve real labelled data traced.

``Training'' also spanned a wider range than gradient descent, and the cheap end of it competed.
\texttt{pjmathematician} (11th, $0.8109$) submits a pipeline that computes no gradient at all, a
zero-shot retriever, a hosted reranker and a closed-form prior, and tunes what the models are asked
instead. A roughly 30-variant sweep over retriever instructions found one whose target-side half
alone was worth $+0.08$ macro on validation, and a failure-driven loop over task-specific reranker
prompts, one to three iterations each accepted only on a fixed candidate pool, carried the pipeline
from $0.806$ to $0.826$. The same team separately trained a 4B agent with GRPO on 243 validation
queries; it finishes below the prompt-optimised pipeline it was meant to replace. Such prompt work
runs through the field's LLM stages generally: hand-written 0--4 rubrics, per-task instructions
rewritten because a job-ad prompt misfits a bare skill mention, and few-shot examples drawn from the
graded validation labels, the same supervision arriving in-context rather than through a loss. Among
accounts documented only by their description, three declare a training-free entry outright:
\texttt{zhangendjb}, whose first submissions fuse stock retrieval channels and use the validation
split only to select modules and fusion weights before a later version adds fine-tuned lanes, and
\texttt{charles2006}, parameter-free on CPU. The distinction the leaderboard rewarded was not
whether a team trained but what it trained \emph{on}.

\paragraph{Taxonomy Usage.} ESCO is a structured taxonomy, carrying a concept hierarchy, skill
types, reuse levels and occupation links. How deeply a system reads it is a question only source
code answers reliably, so the counts here are over the 13 read systems. The flat text fields
dominate: descriptions appear in 12 of the 13 and alternative labels in 9. The \emph{structure} is
the minority case: four systems use hierarchy at inference across the extraction tasks, three touch
skill type or reuse level, and four use no ESCO structure whatever. The descriptions of the other
accounts agree as far as they go, naming alternative labels or aliases twice as often as any
structural field. Where the graph is used the treatment is shallow, mostly hand-crafted count
or diffusion features fed to a gradient-boosted fusion layer. The exception is \texttt{carolzz}
(6th, $0.8132$), the one system to treat ESCO as a graph to search rather than a table to featurise,
seeding a personalized PageRank from its top-64 dense candidates as its own retrieval channel. Even
there the traversal is classical: no system learned a representation over the ESCO graph, in a field
that otherwise reached for every neural component available.

One team found static ESCO-graph features gave a $40\times$ conditional lift on
grade-2 targets but near-zero in-pool precision; the winning entry measured its ESCO attribute
features at $+0.0014$ and switched them off. Under graded relevance the taxonomy paid as a source of
\emph{text} to encode and not as a structure to traverse: 
the flat synonym table was load-bearing while the relational structure,
the part that distinguishes a taxonomy from a list, went largely unused by a field that otherwise
left no cheap gain unclaimed.

The submitted ablations, all on the fixed test set, let the recipe be priced stage by stage.
Stock retrieve-then-rerank with no tuning reaches $0.669$, twelve public dense retrievers
equal-weighted with no training reach $0.806$, within $0.04$ of the winner, task routing is worth
$+0.071$ to one mid-table team, a listwise objective $+0.023$ over pointwise on a fixed retriever,
and switching a pairwise reranker from binary to graded supervision $+0.023$. The
counterexamples bound the recipe from both sides: a 35B-parameter listwise \emph{graded} reranker
on a BM25 retriever gains only $+0.065$ and lands at $0.289$ (reranking cannot repair
retrieval) while stacking ten stock retrievers and rerankers buys $0.005$ over the shipped
baseline (Section~\ref{sec:size}): breadth is not the lever either. Notably, the graded objective
itself was widely \emph{claimed} but rarely load-bearing: among code-shipping teams only four
optimise graded relevance directly, and the measured gains above are real but an order of
magnitude smaller than what retrieval quality and reranking depth are worth.

The grade-composition lens of Figure~\ref{fig:grade-profile} says which family wins which grades.
Lexical and shortcut-prone systems keep most of the grade-4 hits, which are the surface-matchable
ones, and bleed nDCG on the grade-0/1 tail, while the reranked pipelines win by cleaning that tail
(the winner's top-10 holds $0.15$ graded-0 items per query against BM25's $6.1$). What the
leaderboard ultimately ranked is tail hygiene at depth, a quantity binary evaluation leaves
unmeasured.

\subsection{System Sizes}\label{sec:size}

The two trained baselines on the leaderboard are both mpnet-base encoders of $109.5$M parameters:
ConTeXT-match at $0.6075$ and the challenge's CurriculumMatch retriever at $0.6292$. That fixes a
concrete reference point for the question this section asks: what did the field have to spend, in
parameters, to move past it.

\paragraph{What is counted.} For every team that shipped source code we resolved each neural
component of the \emph{inference} pipeline to a Hugging Face repository and read the parameter count
from the Hub's own metadata. We report the base-checkpoint footprint: the sum over the
\emph{distinct} pretrained backbones a pipeline loads, counting a backbone once however many
fine-tuned copies of it the system runs. It is the reproducible quantity, and a floor on the weights
actually held, since these pipelines run several adapted copies of one backbone (the winning entry
fuses thirteen cross-encoder columns that are fine-tunes of five backbones). Models used only for
training, distillation or teacher grading are excluded, verified against each team's own inference
entry point.

\begin{table}[t]
\centering
\footnotesize
\caption{Inference-time size of every submitted system whose code could be read. Params is the
base-checkpoint footprint (distinct pretrained backbones, billions), and $\times$ baseline is the ratio
to CurriculumMatch. $\dagger$ a lower bound: the pipeline also calls a hosted frontier model with no
published weight count. $\ast$ includes at least one checkpoint sized from file bytes rather than
exact metadata.}
\label{tab:size}
\resizebox{\linewidth}{!}{\begin{tabular}{lrrrr}
\toprule
System & nDCG@100 & Backbones & Params (B) & $\times$ baseline \\
\midrule
\texttt{thylinao}\textsuperscript{$\ast$} & 0.8458 & 6 & 13.9 & 127 \\
\texttt{qihang-wang}\textsuperscript{$\ast$} & 0.8323 & 6 & 35.8 & 327 \\
\texttt{yeyang} & 0.8224 & 6 & 45.6 & 417 \\
\texttt{aiden-lee} & 0.8191 & 9 & 243.7 & 2226 \\
\texttt{elara96}\textsuperscript{$\ast$} & 0.8182 & 7 & 53.7 & 490 \\
\texttt{carolzz} & 0.8132 & 7 & 52.2 & 476 \\
\texttt{mingc} & 0.8116 & 4 & 9.2 & 84 \\
\texttt{pjmathematician}\textsuperscript{$\dagger$} & 0.8109 & 2 & 11.6 & 106 \\
\texttt{leonshangguan}\textsuperscript{$\dagger$} & 0.7821 & 2 & 4.6 & 42 \\
\texttt{ccmike} & 0.7746 & 4 & 8.4 & 77 \\
\texttt{zhangguochang} & 0.6342 & 10 & 18.3 & 167 \\
\midrule
\textit{CurriculumMatch} & 0.6292 & 1 & 0.109 & 1 \\
\textit{ConTeXT-match} & 0.6075 & 1 & 0.109 & 1 \\
\textit{BM25} & 0.2250 & 0 & -- & -- \\
\bottomrule
\end{tabular}
}
\end{table}

\begin{figure}[t]
\centering
\includegraphics[width=\linewidth]{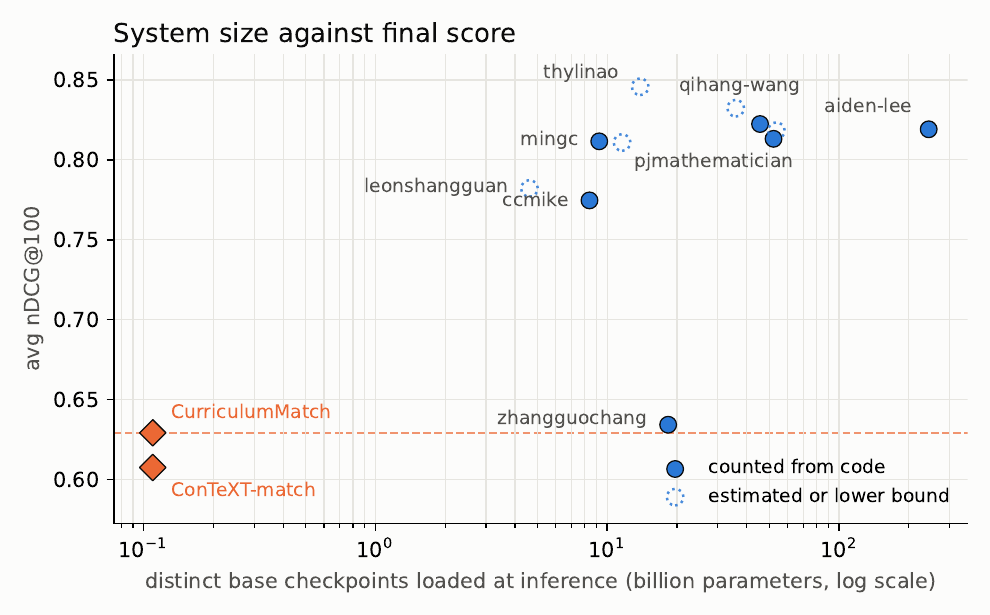}
\caption{Size against score, log scale. Hollow dotted markers are estimates or lower bounds. BM25
($0.2250$, no parameters) cannot be placed on a log axis and is omitted.}
\label{fig:size}
\end{figure}

\paragraph{Nobody competed at the baselines' scale.} Table~\ref{tab:size} and Figure~\ref{fig:size}
show an empty two orders of magnitude. The smallest submitted system we could size is $4.6$B
parameters and the largest is $243.7$B: a span of $42\times$ to $2226\times$ CurriculumMatch, with
no entry anywhere between $109.5$M and $4.6$B. The $0.217$ nDCG separating the top of the
board from the strongest baseline was therefore bought at a minimum of $42\times$ the parameters,
with a median sized system at $167\times$ and the winner itself at $127\times$. The challenge leaves
open what a system at baseline scale could reach, because nobody tried. The one relevant measurement is a participant's own ablation, which puts a zero-shot
Qwen3-Embedding-4B \emph{below} both $109.5$M trained baselines on the validation split
($0.578$ against $0.621$ and $0.642$, macro over the four extraction tasks), and above them only
once task-specific prompts and a reranking cascade are added. Scale alone does not clear the
baselines. Task-specific training and cascade depth do.

\paragraph{Above roughly ten billion parameters, size stops paying.} Across the eleven sized systems
size and score correlate at Spearman $\rho = 0.52$, but that number is carried entirely by the
bottom of the table. Among the eight systems above $0.80$ the footprint spans $26\times$ ($9.2$B
to $243.7$B) and the correlation collapses to $\rho = 0.24$ ($p = 0.57$). The winner, thylinao at
$13.9$B, is $17.6\times$ smaller than the largest system in the field and scores $0.0266$ higher
than it.

\paragraph{The efficient entries.} Two systems are worth naming for beating the state of the art on
something close to a single off-the-shelf model's budget. \texttt{mingc} ($0.8116$, 9th) runs
$9.2$B parameters across four \emph{stock} Qwen3 checkpoints (Embedding-0.6B and Reranker-0.6B
carrying trained LoRA adapters, plus a 4-bit Embedding-4B and a frozen Reranker-4B) with no
hosted model anywhere in the pipeline. That is $84\times$ CurriculumMatch's parameters for $+0.182$ nDCG,
$26\times$ smaller than \texttt{aiden-lee}'s $243.7$B system and $0.0075$ ahead of it, and within
$0.034$ of the winner. \texttt{ccmike} ($0.7746$) is smaller again at $8.4$B and is the one entry
that puts both organiser baselines inside its own ensemble, alongside Qwen3-Embedding-8B and
Qwen3-Reranker-0.6B. At the other end, \texttt{zhangguochang} stacks ten stock retrievers and
rerankers, $18.3$B parameters and $167\times$ the baseline, to finish $0.005$ above CurriculumMatch:
a reminder that ensembling breadth is not the lever either.

\paragraph{What these numbers do not say.} Only a team that shipped code naming its checkpoints can
be sized, so the table covers a little over a quarter of the reported field and is biased towards
teams that documented their work, and the unsized majority is not shown to be smaller or larger. Two
rows are lower bounds because the pipeline calls a hosted frontier model with unpublished weights
(\texttt{pjmathematician} reranks its top-200 with Claude, \texttt{leonshangguan} runs a six-turn
Codex cascade over its top-500), so neither is a small system. Parameter count is also neither
memory nor latency: two counted checkpoints are quantised and two of \texttt{aiden-lee}'s are
mixture-of-experts, which drops its \emph{active} footprint from $243.7$B to roughly $95.6$B. Three
systems include a checkpoint sized from file bytes rather than exact metadata (marked $\ast$), and the
largest such gap, \texttt{deberta-v3-large} at $0.795$B against a true ${\approx}0.44$B, moves no
ordering in the table.

\section{Autoresearch}\label{sec:autoresearch}

Autonomous LLM agents now run machine-learning experiments end to
end~\cite{aiscientist,aicoscientist,mlebench,karpathy2026autoresearch}, and several of the systems
of Section~\ref{sec:methods} already put frontier models inside their pipelines. As an informal
control we let one compete: after the challenge closed, Claude Fable~5~\cite{claudefable} was given
the participant starter kit under participant rules (public data only, ten test evaluations, the
graded test sets and all challenge pages blocked at the harness level) on a single DGX Spark
(119\,GB unified memory), working unattended.

It behaved very much like the field. Starting from the public baselines ($0.64$ validation macro),
it improved by accretion, never subtraction: multi-representation fusion over a dozen embedding
matrices, kNN score propagation, a three-LLM graded-reranking blend, three Qwen3-Reranker scorers,
a 32B pseudo-label grader distilled into a cross-encoder, and finally a 39-feature LambdaMART
trained on the validation qrels, for a cross-validated $0.818$ validation macro. Its one
fine-tuning attempt, on binary-style synthetic supervision, made its retriever worse
($0.642 \to 0.605$) and was dropped in favour of more cascade (the
Section~\ref{sec:graded-vs-binary} failure, rediscovered). The run also
ended the way scale ends: the final test evaluation scored two of the five tasks (Tech $0.786$,
House $0.777$, each just below the ten best per-task scores of Table~\ref{tab:leaderboard}) before
the process died loading one more 32B checkpoint into over-committed memory. The result is
competent, but entirely inside the paradigm Section~\ref{sec:size} and~\cite{} prices: no new
training objective, no small model, a tail bought with parameters until the hardware refused the
next one.
% TODO: add the citation to the paper on autoresearch failures (which also see this kind of failures)

\section{Conclusion}

Decomposing graded relevance into four binary gates made graded annotation objective enough to run
at benchmark scale: an open-weight judge produced 4.3 million dense graded labels, and on a blind
human-refereed manifest the cascade beats the same model asked for the grade directly. Blind
re-annotation shows the labels preserve ordering (quadratic-weighted kappa 0.69--0.73).

That question it served to answer was whether binary evaluation had been hiding the state of the field, and the answer
is yes. The two protocols agree about the broad strokes and disagree about the top: the ordering of
the twelve best teams is uncorrelated with binary MAP ($\tau = 0.06$), the winner under one protocol
places 12th to 16th under the other, and the gap between the best systems and the baselines is two
to three times wider under graded evaluation. The quantity doing the work is one the binary protocol
cannot see: how much explicitly irrelevant material a system tolerates in its ranking window once
the easy positives are placed.

Two things follow, pulling in opposite directions. Semantic skill understanding was not solved
before this challenge: the strongest published skill extractor available off the shelf gives up
$0.217$ nDCG@100 to what a motivated field can reach, a gap binary evaluation reports at a third of
its true size, and participants' own binary-supervised fine-tuning made their systems \emph{worse}
on the graded metric. It is not solved after it either, at least not deployably. The field closed
that tail by brute force: every system we could size runs at least $42\times$, and typically several
hundred times, the parameters of a $109.5$M baseline, with an empty two orders of magnitude in
between and no measurable payoff above roughly $10$B (Section~\ref{sec:size}). Nobody competed at
the baselines' scale, so the challenge cannot say whether a small model trained on graded labels
would have closed the same gap, only that no one tried. An open-source skill recommender an
organisation could run over its own vacancy stream, rather than a thirteen-stage cascade of frontier
checkpoints, is still ahead of us. Beyond skills the methodology transfers directly: any
recommendation task whose ground truth is a handful of positives in a large vocabulary can be graded
this way, and any such task is currently evaluated with a protocol that may, as here, be concealing
how far from solved it is.

\subsection*{Limitations}\label{sec:limitations}

The most consequential limitation is circularity. Our labels come from an LLM cascade, and the
systems that topped the leaderboard are themselves LLM pipelines, several reranking with frontier
models prompted much as our jurors were. Where those models share inductive biases about what makes
a skill relevant to a sentence, the metric rewards agreement with a judge that reasons the way the
ranker does, and no measurement inside this challenge fully separates that from genuine skill
understanding. Two things bound the concern: the human expert, sharing
neither model family nor rubric prompt, reproduces the cascade's ordering at kappa $0.69$, and the
honest reading is that the board partly ranks how well a system's notion of relevance matches a
competent annotator's, which is closer to the deployed objective than counting a handful of
positives. Fully separating the two would take a wholly human-annotated slice, never seen by any
judge, held out as an anchor.

The second limitation is that we measure ranking quality and not who it serves. The framing this
paper opens with, that a skill placed high for the wrong reason attaches a requirement to a role or
a capability to a person, is a fairness claim, and nothing in this challenge tests it. Every number
here is averaged over queries, so a system that keeps its window clean overall while filling it with
nonsense on the subset of vacancies written in a particular register scores exactly like one that
does not. Other work-domain AI-challenges increasingly include this, like TalentCLEF~\cite{gasco2026talentclef},
by scoring paired records that differ only in gender-marked surface elements and reading the
stability of the two rankings. The graded labels make a cheaper version of that available here,
without the paired synthetic corpus a person-matching task needs: because relevance is a property of
the concept a sentence expresses, a meaning-preserving perturbation of the query should leave the
grade profile of the window unchanged, so grade composition at fixed depth can be compared across
perturbed variants of the same sentence.

The third is scope. All annotations are English-only against ESCO v1.1.0, on a taxonomy built to be
multilingual and for a labour market that is not monolingual. Skill recommenders are
deployed against vacancies and profiles written in many languages, often mixed within a single
organisation, and none of the results here speak to whether the tail hygiene that separated this
field survives cross-lingual transfer.

The fourth is that two properties of the released data are payable without doing the task, and both
were paid. Generic ESCO skills (\emph{work in a team}, \emph{solve problems}) are graded 3 or 4 on
10--36\% of extraction queries and almost never 0, so a query-\emph{independent} prior over the
vocabulary carries real nDCG: one participant estimated per-task skill priors from the released
validation labels, validated them leave-one-query-out, and gained $+0.023$ macro without reading the
query at all. The exploit is legitimate, since the labels genuinely do say these skills are usually
recommendable, but it means part of the metric measures the vocabulary's marginal distribution
rather than the query.

The remaining limitations are narrower. The labels carry a documented generous bias on positive
grades, so we recommend graded nDCG as the primary metric and grade $\geq 2$ for any binary cut.
Grade coverage is bounded by the top-2000 prefilter, so relevant skills no candidate generator
surfaces are scored as nonsense. The replacement gate inherits any gaps in the original gold
positives.

Finally, the released validation split is too small to resolve the differences the leaderboard
ranked on. Its 286 queries, of which SkillNorm's 50 contribute $\pm 0.03$ of sampling noise alone,
put the paired minimum detectable effect at 80\% power near $\pm
0.015$~\cite{silchenko2026ranking}, while positions 6 to 11 of Table~\ref{tab:leaderboard} span
$0.003$. The dense middle of the board is unresolved rather than ranked, and only the two gaps this
paper concludes from sit clear of that floor.

\bibliographystyle{elsarticle-num-names}
\bibliography{references}

@inproceedings{veys2026surface,
  title={Surface Matching in Skill Recommenders: Incomplete Evaluation Hides Simplicity Bias},
  author={Veys, Warre and De Lange, Matthias and Decorte, Jens-Joris and Develder, Chris and Demeester, Thomas},
  booktitle={Proceedings of the 20th ACM Conference on Recommender Systems (RecSys '26)},
  year={2026},
  address={Minneapolis, MN, USA},
  publisher={ACM},
  doi={10.1145/3773078.3831842},
  isbn={979-8-4007-2284-4}
}

@inproceedings{delange2026workrb,
    author    = {De Lange, Matthias and Veys, Warre and Retyk, Federico and
                 Deniz, Daniel and Jouanneau, Warren and Zhang, Mike and
                 Bielinski, Aleksander and Jouffroy, Emma and Clobes, Nicole and
                 Baranowska, Nina and Graus, David and Palyart, Marc and
                 Zbib, Rabih and Gkatzia, Dimitra and Demeester, Thomas and
                 De Bie, Tijl and Bogers, Toine and Decorte, Jens-Joris and
                 Van Hautte, Jeroen},
    title     = {{WorkRB}: A Community-Driven Evaluation Framework for {AI} in the Work Domain},
    booktitle = {Proceedings of the 20th ACM Conference on Recommender Systems},
    series    = {RecSys '26},
    year      = {2026},
    publisher = {Association for Computing Machinery},
    address   = {New York, NY, USA},
    location  = {Minneapolis, MN, USA},
    doi       = {10.1145/3773078.3831737},
    url       = {https://doi.org/10.1145/3773078.3831737},
    isbn      = {979-8-4007-2284-4}
  }

@misc{esco2024,
  title={{ESCO}: European Skills, Competences, Qualifications and Occupations (v1.2.0)},
  author={{European Commission}},
  year={2024},
  howpublished={\url{https://esco.ec.europa.eu/en/use-esco/download}},
  note={v1.2.0 released May 2024}
}

@inproceedings{decorte2022negative,
  title={Design of Negative Sampling Strategies for Distantly Supervised Skill Extraction},
  author={Decorte, Jens-Joris and Van Hautte, Jeroen and Deleu, Johannes and Develder, Chris and Demeester, Thomas},
  booktitle={Proc. 2nd Workshop on Recommender Systems for Human Resources (RecSys in HR 2022), co-located with ACM RecSys},
  series={CEUR Workshop Proceedings},
  volume={3218},
  year={2022},
  url={https://ceur-ws.org/Vol-3218/}
}

@article{decorte2025context,
  title={Efficient Text Encoders for Labor Market Analysis},
  author={Decorte, Jens-Joris and Van Hautte, Jeroen and Develder, Chris and Demeester, Thomas},
  journal={IEEE Access},
  volume={13},
  pages={133596--133608},
  year={2025},
  doi={10.1109/ACCESS.2025.3589147}
}

@inproceedings{skillskape,
  title={{JobSkape}: A Framework for Generating Synthetic Job Postings to Enhance Skill Matching},
  author={Magron, Antoine and Dai, Anna and Zhang, Mike and Montariol, Syrielle and Bosselut, Antoine},
  booktitle={Proceedings of the First Workshop on Natural Language Processing for Human Resources (NLP4HR 2024)},
  year={2024},
  pages={43--58},
  address={St. Julian's, Malta},
  publisher={Association for Computational Linguistics},
  url={https://aclanthology.org/2024.nlp4hr-1.4/}
}

@inproceedings{sechidis2011stratification,
  title={On the Stratification of Multi-label Data},
  author={Sechidis, Konstantinos and Tsoumakas, Grigorios and Vlahavas, Ioannis},
  booktitle={Machine Learning and Knowledge Discovery in Databases (ECML PKDD 2011)},
  series={Lecture Notes in Computer Science},
  volume={6913},
  pages={145--158},
  year={2011},
  publisher={Springer},
  doi={10.1007/978-3-642-23808-6_10}
}

@article{voorhees2000variations,
  title={Variations in relevance judgments and the measurement of retrieval effectiveness},
  author={Voorhees, Ellen M.},
  journal={Information Processing \& Management},
  volume={36},
  number={5},
  pages={697--716},
  year={2000},
  publisher={Elsevier},
  doi={10.1016/S0306-4573(00)00010-8}
}

@inproceedings{roitero2018fine,
  title={On Fine-Grained Relevance Scales},
  author={Roitero, Kevin and Maddalena, Eddy and Demartini, Gianluca and Mizzaro, Stefano},
  booktitle={The 41st International ACM SIGIR Conference on Research \& Development in Information Retrieval (SIGIR '18)},
  pages={675--684},
  year={2018},
  publisher={ACM},
  doi={10.1145/3209978.3210052}
}

@misc{silchenko2026ranking,
  title={Thirteen Thousand Skills, in Order},
  author={Silchenko, Maksim},
  year={2026},
  note={System report, team \texttt{thylinao} (solo), RecSys-HR 2026 WorkRB Challenge.
        National University of Singapore. Accessed 1 September 2026},
  howpublished={\url{https://thylinao1.github.io/projects/workrb-skill-ranking.html}}
}

@inproceedings{bgem3,
  title={{M3-Embedding}: Multi-Linguality, Multi-Functionality, Multi-Granularity Text Embeddings Through Self-Knowledge Distillation},
  author={Chen, Jianlyu and Xiao, Shitao and Zhang, Peitian and Luo, Kun and Lian, Defu and Liu, Zheng},
  booktitle={Findings of the Association for Computational Linguistics: ACL 2024},
  pages={2318--2335},
  year={2024},
  publisher={Association for Computational Linguistics},
  url={https://aclanthology.org/2024.findings-acl.137/}
}

@article{qwen3,
  title={Qwen3 Technical Report},
  author={{Qwen Team}},
  journal={arXiv preprint arXiv:2505.09388},
  year={2025},
  url={https://arxiv.org/abs/2505.09388}
}

@misc{qwen3next,
  title={{Qwen3-Next}: Towards Ultimate Training and Inference Efficiency},
  author={{Qwen Team}},
  year={2025},
  howpublished={Qwen Blog},
  note={Released September 11, 2025},
  url={https://qwen.ai/blog?id=4074cca80393150c248e508aa62983f9cb7d27cd}
}

@article{gemma4,
  title={Gemma 4 Technical Report},
  author={{Gemma Team}},
  journal={arXiv preprint arXiv:2607.02770},
  year={2026},
  note={Google DeepMind},
  url={https://arxiv.org/abs/2607.02770}
}

@inproceedings{sentencebert,
  title={{Sentence-BERT}: Sentence Embeddings using {S}iamese {BERT}-Networks},
  author={Reimers, Nils and Gurevych, Iryna},
  booktitle={Proceedings of the 2019 Conference on Empirical Methods in Natural Language Processing and the 9th International Joint Conference on Natural Language Processing (EMNLP-IJCNLP)},
  pages={3982--3992},
  year={2019},
  publisher={Association for Computational Linguistics},
  doi={10.18653/v1/D19-1410}
}

@inproceedings{mpnet,
  title={{MPNet}: Masked and Permuted Pre-training for Language Understanding},
  author={Song, Kaitao and Tan, Xu and Qin, Tao and Lu, Jianfeng and Liu, Tie-Yan},
  booktitle={Advances in Neural Information Processing Systems},
  volume={33},
  pages={16857--16867},
  year={2020}
}

@article{bm25,
  title={The Probabilistic Relevance Framework: {BM25} and Beyond},
  author={Robertson, Stephen and Zaragoza, Hugo},
  journal={Foundations and Trends in Information Retrieval},
  volume={3},
  number={4},
  pages={333--389},
  year={2009},
  publisher={Now Publishers},
  doi={10.1561/1500000019}
}

@misc{claudeopus,
  title={Claude Opus 4.8},
  author={{Anthropic}},
  year={2026},
  howpublished={Model card},
  url={https://www.anthropic.com/news/claude-opus-4-8}
}

@inproceedings{gasco2026talentclef,
  title={Overview of the {TalentCLEF} 2026: Skill and Job Title Intelligence for Human Capital Management},
  author={Gasco, Luis and Fabregat, Hermenegildo and Garc\'ia-Sardi\~na, Laura and Estrella, Paula and Veys, Warre and Carrino, Casimiro Pio and De Lange, Matthias and Deniz Cerpa, Daniel and Rodrigo, \'Alvaro and Decorte, Jens-Joris and Zbib, Rabih},
  booktitle={Working Notes of the Conference and Labs of the Evaluation Forum (CLEF 2026)},
  series={CEUR Workshop Proceedings},
  address={Jena, Germany},
  year={2026},
  publisher={CEUR-WS.org},
  eprint={2606.31692},
  archivePrefix={arXiv},
  primaryClass={cs.CL},
  url={https://arxiv.org/abs/2606.31692}
}

@article{aiscientist,
  title={The {AI} {Scientist}: Towards Fully Automated Open-Ended Scientific Discovery},
  author={Lu, Chris and Lu, Cong and Lange, Robert Tjarko and Foerster, Jakob and Clune, Jeff and Ha, David},
  journal={arXiv preprint arXiv:2408.06292},
  year={2024},
  url={https://arxiv.org/abs/2408.06292}
}

@article{aicoscientist,
  title={Towards an {AI} co-scientist},
  author={Gottweis, Juraj and Weng, Wei-Hung and Daryin, Alexander and Tu, Tao and Palepu, Anil and others},
  journal={arXiv preprint arXiv:2502.18864},
  year={2025},
  note={Google DeepMind},
  url={https://arxiv.org/abs/2502.18864}
}

@article{mlebench,
  title={{MLE-bench}: Evaluating Machine Learning Agents on Machine Learning Engineering},
  author={Chan, Jun Shern and Chowdhury, Neil and Jaffe, Oliver and Aung, James and Sherburn, Dane and Mays, Evan and Starace, Giulio and Liu, Kevin and Maksin, Leon and Patwardhan, Tejal and Weng, Lilian and M{\k{a}}dry, Aleksander},
  journal={arXiv preprint arXiv:2410.07095},
  year={2024},
  url={https://arxiv.org/abs/2410.07095}
}

@misc{karpathy2026autoresearch,
  title={autoresearch: {AI} agents running research overnight},
  author={Karpathy, Andrej},
  year={2026},
  howpublished={\url{https://github.com/karpathy/autoresearch}},
  note={Released March 2026}
}

@misc{claudefable,
  title={Claude Fable 5},
  author={{Anthropic}},
  year={2026},
  howpublished={Model card},
  url={https://www.anthropic.com/news/claude-fable-5-mythos-5}
}

\end{document}